\documentclass[11pt]{article}

\usepackage[preprint]{acl}

\usepackage{times}
\usepackage{latexsym}
\usepackage[T1]{fontenc}
\usepackage[utf8]{inputenc}
\usepackage{microtype}
\usepackage{inconsolata}
\usepackage{graphicx}
\usepackage{booktabs}
\usepackage{multirow}
\usepackage{amsmath,amssymb}
\usepackage{xcolor}
\usepackage{tikz}
\usetikzlibrary{arrows.meta,positioning,calc}
\usepackage{enumitem}
\setlist{nosep,leftmargin=*}

\newcommand{\RQ}[1]{\textbf{RQ#1}}

\title{Late Autumn: LLM Agents Specialize a General\\ Multi-Robot Planner for One Warehouse}

\author{
 \textbf{Yimin Tang\textsuperscript{1}},
 \textbf{Han Zhang\textsuperscript{3}},
 \textbf{Shao-Hung Chan\textsuperscript{3}},
 \textbf{Erdem B{\i}y{\i}k\textsuperscript{1}},
 \textbf{Sven Koenig\textsuperscript{2}},
 \textbf{Jingkai Chen\textsuperscript{3}}
\\
 \textsuperscript{1}Thomas Lord Department of Computer Science, University of Southern California\\
 \textsuperscript{2}Department of Computer Science, University of California Irvine\\
 \textsuperscript{3}Symbotic\\
 \small{\texttt{\{yimintan,biyik\}@usc.edu}, \texttt{svenk@uci.edu}, \texttt{\{hanzhang,shchan,jichen\}@symbotic.com}}
}

\begin{document}
\maketitle

\begin{abstract}
No algorithm is best on all problems, but a real warehouse is a narrow problem: its layout and processes rarely change. Operators still run one general multi-robot planner at many sites, because experts cannot tune it for each one. In this work, which we call Late Autumn, we explore whether LLM coding agents running autonomous research can \emph{specialize} a general planner for one site, trading generality for throughput, much as post-training adapts a pretrained LLM to one task. We study six campaigns on a lifelong planner that combines task assignment with multi-agent path finding, on a floor modeled on an existing automated warehouse. We find that (1)~the agents raise throughput by 22--60\%, and the gains hold on unseen seeds; (2)~they rediscover known MAPF mechanisms, such as corridor-exit priorities, local guidance, and rollouts, and independent campaigns find the same bottleneck; (3)~the gains largely carry over to another layout, a larger fleet, and other service times, shrink with a smaller fleet, and mostly vanish under an outbound-heavy demand mix; and (4)~each final patch (0.7k--2.4k lines) is a chain of scored commits with a median step of 27 lines, so experts can review it, and review finds problems that the score misses; some patches multiply planner compute by over 100$\times$. LLM agents can therefore turn each site's bias into extra throughput at little engineering cost, with reviewable patches; we expect experts to shift toward designing insightful metrics and fast simulators that mirror the real problem.
\end{abstract}

\section{Introduction}
\label{sec:intro}

\begin{figure}[t]
  \centering
  \includegraphics[width=\columnwidth]{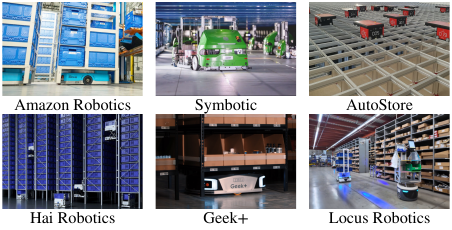}
  \caption{Commercial warehouse robot systems.}
  \label{fig:robots}
\end{figure}

Automated warehouses move goods with fleets of robots that shuttle cases between storage aisles and stations. Commercial systems take many forms (Figure~\ref{fig:robots}): drive units that carry shelves to pickers (Amazon Robotics, Geek+), bots that carry cases through multi-level aisles (Symbotic), robots on a grid above stacked bins (AutoStore), rack-climbing tote robots (Hai Robotics), and robots that work beside human pickers (Locus Robotics). The coordination layer must repeatedly assign tasks to robots and plan collision-free paths for all of them, a lifelong variant of Multi-Agent Path Finding (MAPF) with target assignment~\citep{ma2017lifelong,li2021rhcr,ma2016tapf}. MAPF is NP-hard to solve optimally~\citep{yu2013structure}, so large fleets call for fast suboptimal planners such as PIBT~\citep{okumura2022pibt} and LaCAM~\citep{okumura2023lacam}. We use a lifelong version of ITA-LaCAM~\citep{tang2026italacam}, which couples target assignment with LaCAM search. As a concrete example, we model the floor of one existing automated warehouse; since MAPF models any AGV warehouse as a graph with tasks, the approach is not tied to it.

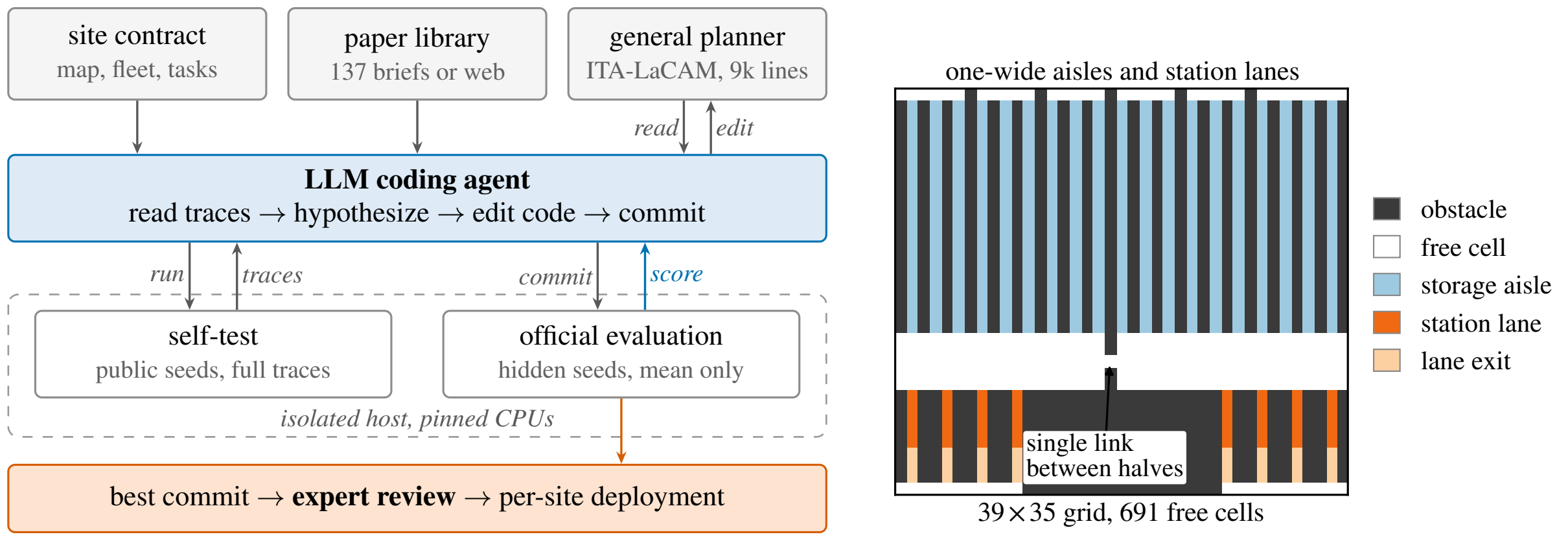
\begin{figure*}[t]
  \centering
  \begin{minipage}[b]{0.535\textwidth}\centering
    \resizebox{\linewidth}{!}{\providecommand{\subtxt}{\fontsize{7.4}{8.6}\selectfont\color{black!62}}
\begin{tikzpicture}[
    font=\fontsize{8.4}{9.8}\selectfont,
    box/.style={draw=#1, rounded corners=2pt, align=center, inner sep=2.4pt, line width=0.6pt},
    box/.default=black!45,
    inbox/.style={box, fill=black!4, minimum width=2.68cm, minimum height=0.95cm},
    evbox/.style={box, fill=white, minimum width=3.7cm, minimum height=0.9cm},
    arr/.style={-{Stealth[length=3.6pt,width=3pt]}, line width=0.65pt, draw=black!65},
    elbl/.style={font=\fontsize{7.4}{8.6}\selectfont\itshape, text=black!65, inner sep=1.2pt},
  ]
  \definecolor{agentc}{HTML}{0072B2}
  \definecolor{agentf}{HTML}{DEEBF7}
  \definecolor{revc}{HTML}{D55E00}
  \definecolor{revf}{HTML}{FDE3D0}
  \node[inbox] (task) at (1.34,0) {site contract\\{\subtxt map, fleet, tasks}};
  \node[inbox] (lib)  at (4.25,0) {paper library\\{\subtxt 137 briefs or web}};
  \node[inbox] (repo) at (7.16,0) {general planner\\{\subtxt ITA-LaCAM, 9k lines}};
  \node[box=agentc, fill=agentf, minimum width=8.5cm, minimum height=0.9cm] (agent) at (4.25,-1.5)
    {\textbf{LLM coding agent}\\read traces $\rightarrow$ hypothesize $\rightarrow$ edit code $\rightarrow$ commit};
  \draw[arr] (task.south) -- (task.south |- agent.north);
  \draw[arr] (lib.south) -- (lib.south |- agent.north);
  \draw[arr] ([xshift=-4pt]repo.south) -- ([xshift=-4pt]repo.south |- agent.north)
    node[elbl, midway, left] {read};
  \draw[arr] ([xshift=4pt]repo.south |- agent.north) -- ([xshift=4pt]repo.south)
    node[elbl, midway, right] {edit};
  \draw[black!40, dashed, rounded corners=3pt, line width=0.5pt] (0.0,-2.5) rectangle (8.5,-3.98);
  \node[evbox] (self) at (2.13,-3.12) {self-test\\{\subtxt public seeds, full traces}};
  \node[evbox] (off)  at (6.37,-3.12) {official evaluation\\{\subtxt hidden seeds, mean only}};
  \node[elbl, anchor=south] at (4.25,-3.97) {isolated host, pinned CPUs};
  \draw[arr] ([xshift=-7pt]self.north |- agent.south) -- ([xshift=-7pt]self.north)
    node[elbl, midway, left] {run};
  \draw[arr] ([xshift=7pt]self.north) -- ([xshift=7pt]self.north |- agent.south)
    node[elbl, midway, right] {traces};
  \draw[arr] ([xshift=-7pt]off.north |- agent.south) -- ([xshift=-7pt]off.north)
    node[elbl, midway, left] {commit};
  \draw[arr, draw=agentc] ([xshift=7pt]off.north) -- ([xshift=7pt]off.north |- agent.south)
    node[elbl, midway, right, text=agentc] {score};
  \node[box=revc, fill=revf, minimum width=8.5cm, minimum height=0.7cm] (rev) at (4.25,-4.62)
    {best commit $\rightarrow$ \textbf{expert review} $\rightarrow$ per-site deployment};
  \draw[arr, draw=revc] (off.south) -- (off.south |- rev.north);
\end{tikzpicture}}\\[2pt]
    {\small (a) auto-research loop}
  \end{minipage}\hfill
  \begin{minipage}[b]{0.43\textwidth}\centering
    \includegraphics[width=\linewidth]{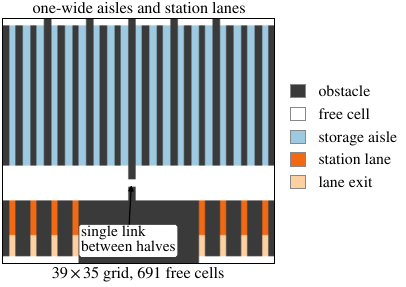}\\[2pt]
    {\small (b) tuned site}
  \end{minipage}
  \caption{(a) The agent edits a general planner for one site, self-tests on public seeds with traces, and is scored only on hidden seeds; experts review the diffs. (b) The tuned site: aisles and station lanes are one-wide corridors, and one cell links the two halves.}
  \label{fig:teaser}
\end{figure*}

This planner is a \emph{general} algorithm, designed for a broad class of layouts and workloads. By the No Free Lunch theorems, averaged over all possible problems, every search algorithm performs equally well~\citep{wolpert1997nfl}. Real sites are far from this average: warehouses differ in aisle geometry, fleet size, carry capacity, demand mix, and service times, but each one runs the same layout and largely fixed processes every day. A planner can therefore trade generality that one site never needs for throughput, much as post-training specializes a pretrained LLM. Until now, only experts could exploit this bias, by inspecting traces and changing cost or priority rules, and no company can do this for its growing number of sites. An agent that can would give each site extra throughput at almost no engineering cost.

LLM-based coding agents now perform parts of this loop~\citep{novikov2025alphaevolve,qu2026coral}, but most of this evidence comes from short programs with cheap, deterministic scores. A warehouse planner differs in three ways. First, the agent edits a complete planner (assignment, search, and simulation) rather than a single heuristic function. Second, the score comes from a stochastic simulation in which many changes are smaller than the noise, so an agent can ``improve'' by chance. Third, a change reaches physical robots only after an engineer understands it and trusts it. We therefore ask:
\begin{enumerate}[label=\textbf{RQ\arabic*},leftmargin=*,widest=4,align=left]
  \item After autonomous research on one site configuration, does the planner's throughput really improve?
  \item What methods do the agents propose?
  \item Are these methods site-specific, or do they work for other warehouse configurations?
  \item How large is each improving change, and can an expert review it to confirm, from first principles, that it is correct?
\end{enumerate}

In this work, which we call Late Autumn, we explore this idea with an existing auto-research system and answer these questions with a case study. An agent receives the planner repository and a task contract that fixes one site configuration. It may change any planner logic, but it is scored only on hidden seeds by an isolated host, and it must commit and evaluate every candidate. We ran six campaigns (three model families, three knowledge settings; 600 official evaluations), re-evaluated each best patch on fresh seeds and on seven held-out configurations, and analyzed every diff that raised the best score. Our results show that every campaign's best patch beats the baseline on all 20 fresh seeds, by 22--63\%, and that the patches keep a significant gain in 40 of 42 held-out configurations. Our contributions are:
\begin{enumerate}
  \item A case study of agent-driven specialization of a lifelong MAPF planner to one site, run with hidden-seed scoring, a mandatory commit-and-evaluate lifecycle, and compute accounting.
  \item Evidence that agents raise throughput by 22--60\% (\RQ{1}).
  \item A catalog of the discovered mechanisms (\RQ{2}), a transfer study (\RQ{3}), and a code-level analysis (\RQ{4}): improving steps are small (median 27 lines), so even the final patches can be reviewed commit by commit, and review finds problems that the score misses.
\end{enumerate}

\section{Problem Definition and Baseline}
\label{sec:problem}

\paragraph{Lifelong TAPF with multi-carry robots.}
Let $G=(V,E)$ be a 4-connected grid graph whose vertices are free cells; self-loops represent waiting. Agents $I=\{1,\dots,N\}$ start at distinct vertices and each can carry at most $K$ tasks. Tasks are released online; a task $\tau=(s_\tau, G_\tau)$ has a pickup vertex $s_\tau$ and a goal set $G_\tau\subseteq V$, any vertex of which completes the delivery. Let $x_i(t)\in V$ be the location of agent $i$ at timestep $t$. A plan is valid if, for all agents $i\neq j$ and timesteps $t$: 1)~$x_i(t{+}1){=}x_i(t)$ or $(x_i(t),x_i(t{+}1))\in E$; 2)~$x_i(t)\neq x_j(t)$ (no vertex conflict); 3)~agents never swap vertices (no edge conflict); and 4)~agent $i$ picks up $\tau$ only by staying at $s_\tau$ for $d_p$ consecutive timesteps while it carries fewer than $K$ tasks, and delivers $\tau$ only by staying at some $g\in G_\tau$ for $d_d$ timesteps. The objective is \emph{throughput}: tasks completed within horizon $H$, divided by $H$. Each planner call has a 1\,s wall-clock limit.

\paragraph{Baseline planner.}
Our planner, a lifelong version of ITA-LaCAM~\citep{tang2026italacam}, replans after every pickup or delivery or when its plan runs out. A Hungarian matching assigns each agent one goal option (a pickup, a goal cell of a carried case, or a costly wait), costed by distance plus service time, with at most one agent per pickup or drop cell; as in ITA-CBS~\citep{tang2023itacbs}, the matching is repaired incrementally at each search node. LaCAM~\citep{okumura2023lacam} with FOCAL node selection then generates configurations with PIBT~\citep{okumura2022pibt} until some agent completes a service. The planner has 9.0k lines of C++ and needs 4.8\,ms per call at the tuned site (\S\ref{sec:setup}): its anytime refinement exits after the first solution (the stopping test compares quantities on different scales), so over 99\% of the budget is unused. It reaches a hidden-seed throughput of 1.23 (pseudo-code in Appendix~\ref{app:baseline}).

\section{Related Work}
\label{sec:related}

\paragraph{MAPF in warehouses.}
PIBT and LaCAM plan for hundreds of agents in milliseconds~\citep{okumura2022pibt,okumura2023lacam,okumura2023engineering}, unlike optimal CBS~\citep{sharon2015cbs,stern2019mapf}. TAPF couples target assignment with path finding~\citep{ma2016tapf,honig2018cbsta,tang2023itacbs,tang2024itaecbs,kumagai2026atp}, and lifelong variants model warehouses with continuous task arrival~\citep{ma2017lifelong,li2021rhcr,chan2024lorr}. Guidance-graph optimization~\citep{zhang2024ggo,zang2024online} and imitation learning~\citep{jiang2024tenthousand} raise throughput; learned planners and fleet foundation models serve many maps~\citep{tang2025railgun,agaskar2025deepfleet}. These methods are designed by researchers for a class of instances; we study whether an agent can do the remaining site-specific engineering.

\paragraph{LLM agents for algorithm engineering.}
LLM-driven program search discovered new constructions and heuristics~\citep{romeraparedes2024funsearch,novikov2025alphaevolve,lange2025shinka,liu2024eoh,ye2024reevo}, agentic systems run research loops over code~\citep{jiang2025aide,lu2024aiscientist,qu2026coral}, and benchmarks evaluate them~\citep{chan2024mlebench,huang2023mlagentbench,jimenez2024swebench,press2025algotune,imajuku2025alebench}. There, programs are small and scores are cheap. LLMs fail when they plan MAPF paths directly~\citep{chen2024llmmapf}; we let them \emph{write planner code} instead, so the planner remains a verifiable search algorithm.

\section{Auto-Research Setup}
\label{sec:setup}

In this section, we describe the auto-research setup: the agent's tools, the task contract it must follow, and how we re-evaluate its results. We use CORAL~\citep{qu2026coral} to run one autonomous coding agent per campaign in its own git worktree (Figure~\ref{fig:teaser}a). The agent can read and edit all code, build, test, inspect traces, and request evaluations. A natural-language \emph{task contract} (Appendix~\ref{app:contract}) sets the rules below.

\paragraph{Site and workload.}
The site the agents tune (Figure~\ref{fig:teaser}b) follows one existing warehouse floor: a $35{\times}39$ grid with 691 free cells, 19 one-wide storage aisles of 20 cells, and 8 one-wide station lanes; a wall splits it into two halves joined by a single cell. An \emph{outbound} task (probability $p_{\text{out}}{=}0.5$) starts at a storage cell and may be delivered to any of 3 cells of one station lane; an \emph{inbound} task starts at a station cell and may be delivered to any of 3 storage cells. New tasks are released so that $2N$ tasks stay unfinished, which keeps the system saturated. We use $N{=}100$, $K{=}2$, $H{=}400$, and $d_p{=}d_d{=}4$.

\paragraph{Fixed site, free algorithm.}
The contract fixes this site and workload. The agent may change any planner, assignment, or goal-selection logic and any algorithm-side default. It may not modify the map, the task process, or the benchmark runner (the evaluation host overwrites them), hard-code seeds or map-specific shortcuts, or fabricate outputs; the grader recomputes throughput from the completed tasks.

\paragraph{Isolated evaluation.}
Throughput depends on how much search fits into 1\,s, so builds and benchmarks run on a separate host with pinned 10-core slots. A \emph{self-test} runs public seeds 0--9 and returns per-seed traces. An \emph{official evaluation} runs 10 hidden seeds and returns only aggregates. The same binary is reproducible on one slot, but contention can move the 10-seed mean by about 0.025; six evaluations of the unmodified baseline spanned 1.198--1.249. The contract therefore treats mean changes below 0.03 as noise.

\paragraph{Commit-and-evaluate lifecycle.}
Every candidate that changes behavior, builds, and yields valid output must be committed and officially evaluated \emph{before} the agent keeps, reverts, or abandons it. This rule prevents silent cherry-picking and gives us the complete record. CORAL adds periodic prompts to \emph{reflect} (every 2 evaluations), \emph{consolidate} (every 8), and \emph{pivot} (on plateaus). A campaign stops after 100 official evaluations.

\paragraph{Safety property.}
The simulator validates every executed joint move for vertex and edge conflicts. As long as a patch leaves this validator unchanged, which review confirmed for all 102 new-best steps, a patch can lower throughput but cannot produce a collision in an executed plan.

\paragraph{Knowledge settings and campaigns.}
In the \emph{paper-grounded} setting, every idea must cite one of 137 briefs of MAPF-related papers and competition reports. In the \emph{open} setting, any source is allowed and must be named; \emph{open+web} also enables web search. We ran six campaigns from the same commit at high reasoning effort, with four models in three agent harnesses (Table~\ref{tab:campaigns}).

\paragraph{Re-evaluation and transfer.}
A campaign's best hidden score is a maximum over noisy evaluations. We therefore rebuilt the baseline and each best commit and ran them on 20 fresh seeds for the tuned site and for seven held-out configurations, each changing one factor: an alternative layout (Appendix~\ref{app:layout}), 60 or 150 robots, $K{=}1$, an 80/20 outbound/inbound mix, and service durations of 2 or 8, a total of 1,120 episodes (7 builds, 8 configurations, 20 seeds). \emph{Gain} is the paired throughput difference to the baseline on the same seeds, with a 95\% CI over seeds; a \emph{new-best step} is an official evaluation that raises a campaign's best hidden score; its \emph{size} is the number of changed lines of algorithm code, excluding tests, tools, and build files. All builds use \mbox{GCC~10.5 \texttt{-O3};} the re-evaluation runs used a 32-core AMD EPYC~9R14 host with 64\,GB RAM.

\section{Results}
\label{sec:results}

\begin{table*}[t]
\centering
\small
\setlength{\tabcolsep}{4.2pt}
\begin{tabular}{lllrrrrrrrr}
\toprule
Model & Harness & Ideas & Evals & Hours & \multicolumn{2}{c}{Hidden best} & \multicolumn{2}{c}{Fresh seeds} & Evals to & Planner s \\
 & & & & & TP & gain & TP & gain & 90\% gain & / episode \\
\midrule
\multicolumn{3}{l}{Unmodified baseline (6 evals)} & -- & -- & 1.230 & -- & 1.209 & -- & -- & 1.8--3.8 \\
Opus 5.5 & Claude Code & paper & 100 & 7 & 1.973 & +60\% & 1.968 & +63\% & 24 & 21.2 \\
Opus 5.5 & Claude Code & open & 100 & 12 & 1.875 & +52\% & 1.869 & +55\% & 73 & 189.9 \\
Fable 5.1 & Kiro & paper & 100 & 8 & 1.884 & +53\% & 1.861 & +54\% & 67 & 11.3 \\
Fable 5.1 & Kiro & open+web$^\dagger$ & 100 & 9 & 1.758 & +43\% & 1.746 & +44\% & 27 & 94.1 \\
GPT-6 Luna & Codex & paper & 100 & 22 & 1.500 & +22\% & 1.476 & +22\% & 83 & 2.9 \\
GPT-6.1 Sol & Codex & paper & 100 & 34 & 1.498 & +22\% & 1.497 & +24\% & 49 & 342.1 \\
\bottomrule
\end{tabular}
\caption{The six campaigns. Hidden-best gain: relative to the baseline mean 1.230. Fresh seeds: best commit on 20 unseen seeds. Hours: active run time. Planner s: per 400-step episode (baseline: range over six evaluations). $^\dagger$Web search enabled but never used.}
\label{tab:campaigns}
\end{table*}

In our experiments, we evaluate the best patch of each campaign against the unmodified baseline and answer the four research questions in turn.

\subsection{RQ1: Does throughput really improve?}

\begin{figure}[t]
  \centering
  \includegraphics[width=\columnwidth]{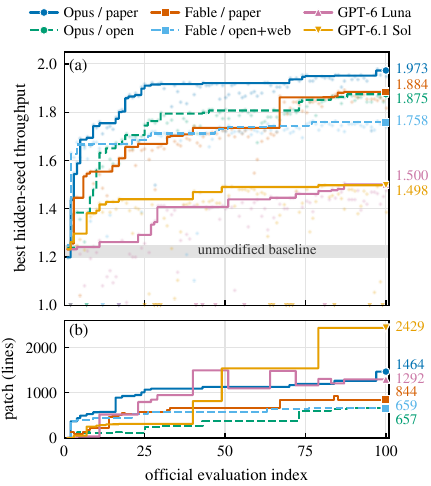}
  \caption{(a) Best-so-far hidden-seed throughput (dots: single evaluations, those below 1.0 on the bottom edge; gray band: six baseline evaluations). (b) Size of the best commit so far (changed lines of algorithm code); it drops when a new best comes from a branch with less code. Solid: paper-grounded; dashed: open.}
  \label{fig:trajectory}
\end{figure}

\begin{figure*}[t]
  \centering
  \includegraphics[width=\textwidth]{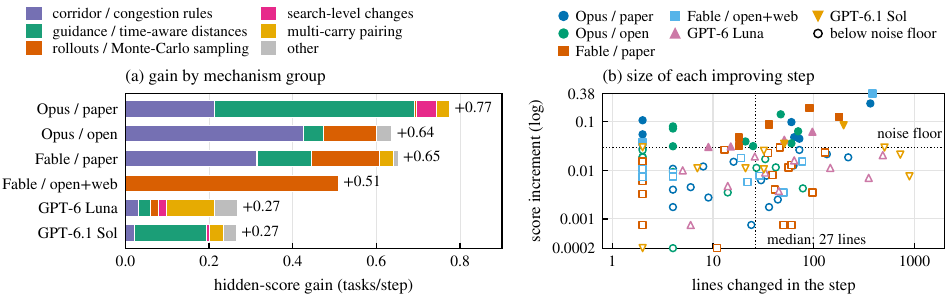}
  \caption{(a) Hidden-score gain per campaign by mechanism group; each new-best step counts toward its main change, and retunes toward the mechanism they tune. (b) Size of each new-best step versus its increment; filled marks exceed the 0.03 noise floor. Sizes count algorithm code only.}
  \label{fig:analysis}
\end{figure*}

\paragraph{Yes, by 22--60\%.}
Every campaign raised hidden-seed throughput far beyond the noise floor (Table~\ref{tab:campaigns}, Figure~\ref{fig:trajectory}). The best campaign (Opus, paper-grounded) reached 1.973 tasks per timestep, 60\% above the baseline mean of 1.230. The four Claude-family campaigns gained 43--60\%, and both GPT campaigns plateaued near 1.50 (+22\%).

\paragraph{The gains hold on fresh seeds.}
On 20 seeds that no agent had seen, every best patch scores within 0.001--0.024 of its hidden best (Table~\ref{tab:campaigns}), so selection bias is small, and beats the baseline on all 20 seeds; for example, the Opus/paper patch reaches 1.968 (+63\%).

\paragraph{Gains come early, partly from compute.}
The Claude-family campaigns reached 65--86\% of their final gain within 20 evaluations (0.6--1.9 hours); only 27 of the 102 new-best steps exceed the 0.03 noise floor. Relative to their baseline evaluation, the paper-grounded Claude patches use 4--10$\times$ more planner time and the open-setting patches 53--91$\times$ (mean calls of 0.3--0.6\,s against the 1\,s limit). GPT-6.1 Sol's last new best (+0.008) raised its planner time 87-fold (112$\times$ its baseline).

\begin{table}[t]
\centering
\small
\setlength{\tabcolsep}{2.2pt}
\begin{tabular}{@{}lcccccc@{}}
\toprule
Mechanism (pipeline stage) & OP & OO & FP & FW & GL & GS \\
\midrule
Corridor-exit priority (PIBT) & .10 & .14 & .09 & -- & .03$^\ast$ & -- \\
Corridor-blocking cost (assign.) & .12 & .28 & .20 & $\times$ & -- & -- \\
Local guidance (PIBT order) & .35 & .04 & -- & -- & $\times$ & $\times$ \\
Flow-aware distances / routes & -- & .00 & .13 & -- & .03 & .17 \\
Time-aware assignment & .13 & -- & -- & -- & $\times$ & -- \\
Rollouts / Monte-Carlo PIBT & .00 & .12 & .16 & .51 & .02 & -- \\
$K{=}2$ insertion / pairing & .03 & -- & .03 & -- & .11 & .04 \\
Search-level changes & .05 & -- & -- & -- & .02 & .01 \\
\bottomrule
\end{tabular}
\caption{Gain of the new-best steps that introduced or tuned each recurring mechanism ($\times$: tried, never a new best). OP/OO: Opus paper/open; FP/FW: Fable paper/open+web; GL/GS: GPT-6 Luna/6.1 Sol. $^\ast$Around articulation cells only.}
\label{tab:mechanisms}
\end{table}

\subsection{RQ2: What did the agents propose?}

We read all 102 diffs that raised the best score, together with the agents' notes (Table~\ref{tab:mechanisms}, Figure~\ref{fig:analysis}a).

\paragraph{Independent campaigns found the same bottleneck.}
On this floor, 470 of the 691 free cells lie in one-wide corridors, and a servicing robot blocks everyone behind it. Three Claude campaigns found this, from traces (``corridor entrances are occupied 89\% of the time'') or from a cited paper, and converged on two rules. A \emph{corridor-exit priority} gives a PIBT priority boost to a robot inside a corridor whose goal lies outside it, so that priority inheritance pushes entering robots back instead of pushing the leaving robot deeper (+0.09 to +0.14; cf.\ PIBTTP~\citep{fujitani2022pibttp}). A \emph{corridor-blocking cost} makes pickup and drop cells behind occupied corridor cells more expensive in the matching (+0.12 to +0.28; cf.\ idle-time-aware assignment~\citep{kou2019ito}). All three detected corridors generically, as components of degree-2 vertices. These rules provide 28--67\% of those campaigns' gains.

\paragraph{Agents spent the idle budget.}
Agents ported local guidance~\citep{arita2025lg,arita2026lllg} (+0.35 in Opus/paper), added flow-dependent edge weights~\citep{zhang2024ggo,zang2024online} (+0.13 in Fable/paper), and built rollout planners; Fable/open+web replaced LaCAM's search with the best of many noisy PIBT rollouts (+0.38 in one step).

\paragraph{Sources and failures.}
Paper-grounded campaigns cited a brief in 97--99\% of evaluations but used only 18--44 of 137 briefs; 79\% of Opus/open candidates came from its own diagnosis; the web was never used. Comparing lineages, we find two reasons why the GPT campaigns plateaued near 1.50. First, they turned to the library instead of the traces and never built corridor rules for the aisles; the same local-guidance paper gave Opus +0.24 but GPT 0.98--1.30, because GPT used a 5-step window and no warm start. Second, they followed CORAL's ``hardest idea, three evaluations'' prompt literally, with median attempts of 66 and 290 added lines (Claude: 4--16).

\subsection{RQ3: Specific or general?}

\begin{figure}[t]
  \centering
  \includegraphics[width=\columnwidth]{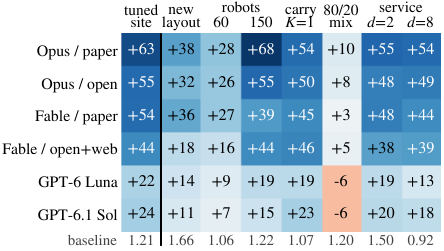}
  \caption{Paired gain (\%) of each unchanged best patch over the baseline, 20 fresh seeds per cell; all 95\% CIs exclude zero. Bottom row: baseline throughput.}
  \label{fig:transfer}
\end{figure}

\paragraph{Code-level generality.}
We graded each frontier step as G0 (generic features only), G1 (generic code, layout-tuned constants), or G2 (scenario facts). Of 77 Claude-family steps, 12 are G0, 59 G1, and 6 G2; none hard-codes seeds, coordinates, or the map.

\paragraph{Transfer.}
Applied unchanged to seven held-out configurations (Figure~\ref{fig:transfer}), the patches gain significantly in 40 of 42 cells. On the alternative layout, the patches keep 57--91\% of their absolute gain, because its aisles are still one-wide. We observe a clear trend with congestion: with 60 robots only 25--43\% of the absolute gain remains, whereas with 150 robots 64--110\% remains. This is because the discovered mechanisms target blocking in one-wide corridors, which grows with fleet size. Relative gains also stay close to the tuned ones for single-carry robots and service durations of 2 and 8. The exception is the 80/20 outbound-heavy mix, which sends most deliveries to the 8 station lanes: the Claude patches keep only +3 to +10\%, and both GPT patches \emph{lose} 6\%. Site-specific campaigns thus produced mostly general improvements, but a new demand mix calls for a new campaign.

\subsection{RQ4: How large are the changes, and can humans review them?}

\paragraph{Steps and patches are small.}
The median improving step changes 27 lines, 27 of the 102 steps only retune a constant, and the largest gains came from 36--383-line mechanisms (Figures~\ref{fig:trajectory}b, \ref{fig:analysis}b; Appendix~\ref{app:example}). The final patches (0.7k--1.5k lines for Claude, 1.3k--2.4k for GPT) are modest next to the 9k-line planner, and each is a chain of 13--49 officially scored commits (median 4--51 lines) that an expert can review one at a time; this also exposes dead code (26\% of the lines added by Opus/paper).

\paragraph{Review is feasible and necessary.}
Of the 24 frontier steps of Opus/paper, 7 are constant retunes, 10 are local cost or priority rules that an expert can verify from first principles, and 7 need careful review. Review found problems that the score cannot reveal: (i)~the Opus/open rollout stops at 880\,ms and then starts the main solve with a \emph{fresh} 1\,s deadline, so the per-call limit holds only empirically; (ii)~a Monte-Carlo step redraws PIBT's random tie-breaks, so the executed configuration can differ from the scored one; and (iii)~dwell-sized constants and $K{=}2$ formulas are hard-coded.

\section{Lessons and Conclusion}
\label{sec:lessons}
\label{sec:conclusion}

LLM agents can specialize a general algorithm to one site for extra performance: campaigns raised throughput by 22--60\% with trace-diagnosed corridor rules and with guidance and rollouts that use idle compute, and the gains mostly transferred. As a site's layout and processes rarely change, this turns site bias into nearly free throughput and lower logistics costs. The final patches can be reviewed commit by commit. Since agents optimize exactly what the score and simulator measure, we expect experts to shift from hand-tuning algorithms toward designing insightful metrics and fast simulators that mirror the real problem. We recommend short campaigns (about 20 evaluations), unseen-seed checks, and a graded call-time limit. Next, we plan to deploy reviewed patches on physical robots.

\section*{Limitations}
\label{sec:limitations}
This is a case study of one planner and one family of sites. All results come from simulation; we have not yet deployed an agent-generated patch on physical robots. We ran one campaign per (model, harness, knowledge setting) combination, so differences between campaigns combine the effects of the model, the agent harness (Claude Code, Kiro, Codex), the knowledge setting, and run-to-run variance; the comparison is indicative, not causal. The best hidden score of each campaign is selected from up to 100 noisy evaluations and is therefore optimistic; we report fresh-seed re-evaluations to correct for this. Our reviewability analysis is based on diff statistics and on an LLM-assisted code reading whose key findings we checked against the source. Finally, the agents optimized throughput only. They could trade other operational metrics, such as task latency, energy use, or robot wear, which we did not measure.

\section*{Ethical Considerations}
Agent-generated changes to software that controls physical robots carry safety risk. In our protocol, agents never act on hardware; they edit planner code that is scored in simulation, and the planner's collision checks remain the final safety layer. We argue that every patch should pass expert review and staged testing before deployment. The intended use is to let a small engineering team support many sites; this may change the work of operations engineers, and we recommend keeping humans responsible for deployment decisions. Higher throughput from the same robots can lower the cost of warehouse logistics for operators and their customers; this benefit should be weighed against the extra planner compute that some patches need. No personal data was used.

\bibliography{custom}

\appendix
\section{Baseline Planner Pseudo-Code}
\label{app:baseline}

\begin{figure}[h]
\centering
\resizebox{\columnwidth}{!}{%
\begin{tabular}{@{}r@{\hspace{0.6em}}l@{}}
\multicolumn{2}{@{}l}{\textbf{Input:} grid $G$, $N$ agents, capacity $K$, durations $d_p,d_d$, horizon $H$}\\
1 & $x\leftarrow$ distinct random cells; release $2N$ tasks; plan $\leftarrow\emptyset$\\
2 & \textbf{for} $t=0,\dots,H-1$ \textbf{do}\\
3 & \quad\textbf{if} $t{=}0$, a service finished, or plan is empty \textbf{then}\\
4 & \qquad reset assigned-but-unpicked tasks to pending\\
5 & \qquad $\Omega_i\leftarrow$ pickups (if $|P_i|{<}K$), goals of carried $P_i$, wait\\
6 & \qquad $c_{i\omega}\leftarrow(\mathrm{dist}(x_i,\omega)+d_\omega)\,\sigma_{i\omega}+\text{offset}_{i\omega}$\\
7 & \qquad $A_0\leftarrow$ Hungarian$(c)$, one agent per pickup/drop cell\\
8 & \qquad plan $\leftarrow$ LaCAM-FOCAL$(x,A_0)$ with PIBT successors and\\
  & \qquad\quad incremental assignment repair, until a service ends\\
9 & \quad $x\leftarrow$ next configuration of plan (no vertex/swap conflicts)\\
10 & \quad update dwell counters; pick up or deliver when counter $\ge d$\\
11 & \quad release tasks to keep $2N$ unfinished\\
12 & \textbf{return} \#completed$/H$
\end{tabular}}
\caption{Baseline lifelong ITA-LaCAM loop.}
\end{figure}

\section{Held-Out Layout}
\label{app:layout}
\begin{figure}[h]
  \centering
  \includegraphics[width=0.36\columnwidth]{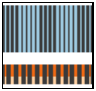}
  \caption{The alternative layout used in the transfer study: $37{\times}39$ cells, 835 free cells, an open top corridor, no dividing wall, and 13 station lanes; colors as in Figure~\ref{fig:teaser}b.}
\end{figure}

\section{Task Contract Excerpt}
\label{app:contract}
The contract was given to the agents in Chinese; we translate its key rules. \emph{``You may generally change planner, assignment, or goal-candidate behavior. You may not change the fixed official target, hard-code seed/task/map shortcuts, fabricate CSV, trace, or schedule outputs, or optimize only the output format.''} \emph{``A mean change smaller than 0.03 on public seeds is not a signal; prefer trace-derived behavior signals.''} \emph{``Once a behavior-changing candidate is finalized, builds, and produces valid output, you must commit it and run an official evaluation, whether your self-test shows improvement, no change, or regression.''} \emph{``Iteration time is also a resource \ldots{} only accept a clearly higher runtime cost when the throughput evidence is strong (about 0.1 or more).''}

\section{Per-Campaign Frontier Details}
\label{app:frontier}
Table~\ref{tab:frontier_detail} summarizes the frontier of each campaign: the number of new-best steps, how many exceed the noise floor, the number of single-constant retunes, the generality grades, and the final patch size. The full per-step table (campaign, evaluation index, commit, hidden score, increment, lines added and deleted, mechanism group, mechanism, and the agent's evaluation message) is the ancillary file \texttt{frontier\_steps.csv}.

\begin{table}[h]
\centering
\footnotesize
\setlength{\tabcolsep}{3pt}
\resizebox{\columnwidth}{!}{%
\begin{tabular}{@{}lrrrcr@{}}
\toprule
Campaign & Steps & $>$.03 & Retunes & G0/G1/G2 & Patch \\
\midrule
Opus / paper & 24 & 6 & 7 & 5/15/4 & +1454/$-$10 \\
Opus / open & 20 & 8 & 10 & 3/17/0 & +635/$-$22 \\
Fable / paper & 21 & 5 & 4 & 3/16/2 & +824/$-$20 \\
Fable / open+web & 12 & 2 & 6 & 1/11/0 & +558/$-$101 \\
GPT-6 Luna / paper & 14 & 4 & 0 & 12/2/0 & +1264/$-$28 \\
GPT-6.1 Sol / paper & 11 & 2 & 0 & 11/0/0 & +2372/$-$57 \\
\bottomrule
\end{tabular}}
\caption{Frontier statistics per campaign. Retunes: steps that only change constants. Patch: lines of algorithm code added/deleted from the baseline to the best commit (tests, tools, and build files excluded).}
\label{tab:frontier_detail}
\end{table}

\section{Example of an Improving Step}
\label{app:example}
Figure~\ref{fig:example} shows the C++ code added by the second evaluation of the Opus/open campaign (commit \texttt{3400f25f}): corridor detection and the corridor-exit priority. The step adds 45 lines of algorithm code, removes 2, and raised the hidden score from 1.238 to 1.380. A reviewer can check its two claims, that corridors are connected components of vertices with degree at most 2 and that only unsatisfied agents whose goal lies outside their corridor are boosted, directly from the code.

\begin{figure*}[t]
\footnotesize
\begin{verbatim}
// (in the TAPFPlanner constructor)
auto next_component = 0;
for (auto v : ins->G.V) {
  if (v->neighbor.size() > 2 || corridor_component[v->id] >= 0) continue;
  auto stack = std::vector<Vertex*>{v};
  corridor_component[v->id] = next_component;
  while (!stack.empty()) {
    auto u = stack.back(); stack.pop_back();
    for (auto w : u->neighbor) {
      if (w->neighbor.size() > 2 || corridor_component[w->id] >= 0) continue;
      corridor_component[w->id] = next_component; stack.push_back(w);
    }
  }
  ++next_component;
}
// (at the LaCAM root)
// An agent inside a single-lane corridor whose goal lies outside it is
// leaving; give it precedence so agents entering the corridor do not push
// it deeper.
constexpr float kCorridorExitPriority = 10.0f;
auto boosted = false;
for (auto i = 0; i < N; ++i) {
  if (S_init->satisfied[i] || S_init->service_assignment[i] >= 0) continue;
  const auto comp = corridor_component[S_init->C[i]->id];
  if (comp < 0) continue;
  const auto goal = assigned_goal(S_init->assignment, i);
  if (goal == nullptr || corridor_component[goal->id] == comp) continue;
  S_init->priorities[i] += kCorridorExitPriority; boosted = true;
}
if (boosted) {
  std::sort(S_init->order.begin(), S_init->order.end(), [&](int a, int b) {
    return S_init->priorities[a] > S_init->priorities[b]; });
}
\end{verbatim}
\caption{An agent-written improving step (corridor-exit priority, +0.142 hidden throughput), from the diff of commit \texttt{3400f25f}. Code and comments are verbatim; we merged some short lines, dropped enclosing braces, and added the two location notes in parentheses.}
\label{fig:example}
\end{figure*}

\section{Photo Credits}
\label{app:credits}
Figure~\ref{fig:robots}: AutoStore robots, photo by Euku, Wikimedia Commons, CC BY-SA 4.0; Amazon Robotics, Symbotic, Hai Robotics, Geek+, and Locus Robotics, product images from the companies' websites. Images were cropped.

\end{document}